\documentclass[conference]{IEEEtran}
\IEEEoverridecommandlockouts
\usepackage{cite}
\usepackage[T1]{fontenc}
\usepackage[utf8]{inputenc}
\usepackage{amsmath,amsthm,mathtools,amssymb}
\usepackage{algorithmic}
\usepackage{graphicx}
\usepackage{textcomp}
\usepackage[normalem]{ulem}
\usepackage{url}

\usepackage[usenames,dvipsnames,svgnames,table]{xcolor}
\definecolor{blue}{rgb}{0,0,1}

\definecolor{dgreen}{rgb}{0.0,0.5,0.0}

\definecolor{lightBlue}{rgb}{0.,0.5,0.5}

\begin{document}

\title{Exploring deterministic frequency deviations with explainable AI
\thanks{We gratefully acknowledge support from the German Federal Ministry of Education and Research (BMBF grant no. 03EK3055B) and the Helmholtz Association via the \textit{Helmholtz School for Data Science in Life, Earth and Energy} (HDS-LEE). This project has received funding from the European Union’s Horizon 2020 research and innovation programme under the Marie Sk\l{}odowska-Curie grant agreement No. 840825.}
}

\author{\IEEEauthorblockN{Johannes Kruse\IEEEauthorrefmark{1}\IEEEauthorrefmark{2}\IEEEauthorrefmark{4}, Benjamin Sch\"afer\IEEEauthorrefmark{3}\IEEEauthorrefmark{5}, Dirk Witthaut\IEEEauthorrefmark{1}\IEEEauthorrefmark{2}} 
\IEEEauthorblockA{\IEEEauthorrefmark{1}Forschungszentrum J\"ulich, Institute for Energy and Climate Research - Systems Analysis and Technology Evaluation \\ (IEK-STE),  52428 J\"ulich, Germany } 
\IEEEauthorblockA{\IEEEauthorrefmark{2}Institute for Theoretical Physics, University of Cologne, 50937 K\"oln, Germany}
\IEEEauthorblockA{\IEEEauthorrefmark{3}School of Mathematical Sciences, Queen Mary University of London, London E1 4NS, United Kingdom}
\IEEEauthorblockA{\IEEEauthorrefmark{5}Faculty of Science and Technology, Norwegian University of Life Sciences, 1432 Ås, Norway}
\IEEEauthorblockA{\IEEEauthorrefmark{4} Email: jo.kruse@fz-juelich.de }}

\maketitle

\begin{abstract}
Deterministic frequency deviations (DFDs) critically affect power grid frequency quality and power system stability. A better understanding of these events is urgently needed as frequency deviations have been growing in the European grid in recent years. DFDs are partially explained by the rapid adjustment of power generation following the intervals of electricity trading, but this intuitive picture fails especially before and around noonday. In this article, we provide a detailed analysis of DFDs and their relation to external features using methods from explainable Artificial Intelligence. We establish a machine learning model that well describes the daily cycle of DFDs and elucidate key interdependencies using SHapley Additive exPlanations (SHAP). Thereby, we identify solar ramps as critical to explain patterns in the Rate of Change of Frequency (RoCoF).
\end{abstract}

\section{Introduction}
The balance of power generation and demand is central for the stability of our power grids. The grid frequency reflects this balance, since an overproduction of power leads to a rise and an undersupply to a drop of the frequency \cite{machowskiPowerSystemDynamics2008}. 
The need for power balancing therefore translates into constraints on the grid frequency to limit large deviations from the set point of 50/60 Hz through adequate control measures. This is critical as large frequency deviations, such as deviations of more than $200$ mHz in Continental Europe, can trigger the disconnection of loads with severe consequences for the power consumers.

An important threat to frequency stability results from deterministic frequency deviation (DFDs). DFDs occur regularly at the beginning of hourly or sub-hourly intervals and can be observed in various large-scale power grids such as the Continental European \cite{schaferNonGaussianPowerGrid2018}, the Great Britain \cite{homanGridFrequencyVolatility2021} and the Nordic synchronous areas \cite{liFrequencyDeviationsGeneration2011}. Large DFDs lead to a depletion of frequency control reserves at the beginning of hourly or sub-hourly intervals, thus making the system vulnerable to additional, unforeseen disturbances or failures \cite{entso-eaisblReportDeterministicFrequency2019}. For example, the combination of a DFD and a measurement failure led to an extreme deviation of nearly 200 mHz on the 10th of January 2019, in Continental Europe.  In this area, the number of large frequency deviations has increased constantly thus requiring an intensified control of large DFDs \cite{entso-eaisblReportDeterministicFrequency2019}.

A common model explains DFDs through the effect of scheduled changes in the power generation due to block-wise electricity trading \cite{weissbachHighFrequencyDeviations2009}. In this view, the direction of DFDs during the day is often related to the load ramp, i.e. to the slope of the load curve \cite{weissbachHighFrequencyDeviations2009, liFrequencyDeviationsGeneration2011, gorjaoDatadrivenModelPowergrid2019} and a connection between large frequency events and the load ramp is suggested \cite{homanGridFrequencyVolatility2021,perssonFrequencyEvaluationNordic2017}. However, not only load ramps but many other external features such as electricity prices or forecasting errors affect power grid frequency deviations \cite{kruseRevealingRisks2021}.

Methods from Machine Learning (ML) are excellent candidates to model the complex effects of multiple features on power grid frequency fluctuations. Due to control measures, frequency dynamics exhibit complex non-linear dependencies \cite{machowskiPowerSystemDynamics2008} and measurement errors further modify many publicly available grid frequency measurements \cite{krusePredictabilityPowerGrid2020}. Moreover, drivers of frequency deviations such as load or generation ramps are strongly correlated \cite{kruseRevealingRisks2021}. Modern ML methods, such as Gradient Tree Boosting, are able to extract non-linear dependencies even from noisy data with correlated features \cite{hastieElementsStatisticalLearning2016}, which is not possible in a simple correlation analysis. They can further harness the growing amount of power system data, which has been made publicly available in the past years \cite{hirthENTSOETransparencyPlatform2018, rydingorjaoOpenDatabaseAnalysis2020}. However, complex ML models are often black-boxes, which impede a scientific understanding of the model structure \cite{roscherExplainableMachineLearning2020}.

Tools from eXplainable Artificial Intelligence (XAI) enable us to understand and visualise the dependencies captured by the ML model \cite{barredoarrietaExplainableArtificialIntelligence2020}. In particular, the recently introduced SHapely Additive exPlanations (SHAP) values offer a numerically efficient way to quantify the impact of different features on the model output \cite{lundbergLocalExplanationsGlobal2020a}. Within the set of methods to measure feature effects, SHAP values guarantee certain optimal properties and avoid inconsistencies within other approaches \cite{lundbergUnifiedApproachInterpreting2017, lundbergConsistentIndividualizedFeature2019}. Combining non-linear ML models and SHAP values thus enables us to examine the effect of multiple external features on DFDs in a consistent way.

Here, we explore hourly DFDs in the Continental European (CE) synchronous area with an explainable ML model introduced in ref. \cite{kruseRevealingRisks2021}. In sec. \ref{sec:methods}, we introduce the ML model, as well as the common load-based model of DFDs, which is based on the effect scheduled-based generation. In sec. \ref{sec:results}, we first evaluate how well the load-based DFD model reproduces the daily pattern of DFDs in CE. Then, we explore the deficits of the load-based model with SHAP values and finally discuss a refined physical view of DFDs, which integrates the insights from our explainable ML.

\section{Methods}
\label{sec:methods}
\subsection{Modelling deterministic frequency deviations}
DFDs are regular deviations of the grid frequency that occur at the beginning of hourly or sub-hourly intervals. For instance, the grid frequency sample from an evening hour in CE (fig.~\ref{fig:1}) exhibits a steep slope at the full hour and after 30 min. These large slopes even persist after averaging over multiple days, which demonstrates their deterministic nature \cite{weissbachHighFrequencyDeviations2009}. In particular, the direction of the slope shows a regular daily profile. In the morning, the deviations typically point upward and in the evening we typically observe negative slopes.  

We quantify these DFDs with the Rate of Change of Frequency (RoCoF) at the beginning of the hour (fig. \ref{fig:1}). The hourly DFDs dominate the daily profile of the grid frequency \cite{schaferNonGaussianPowerGrid2018}. Moreover, the hourly interval is still the most important time scale for power system operation in CE, since electricity is mostly traded on an hourly basis  \cite{weissbachImpactCurrentMarket2018}. Therefore, we focus on the hourly RoCoF to model DFDs in this study.

\begin{figure*}
    \centering
    \includegraphics[width=\textwidth]{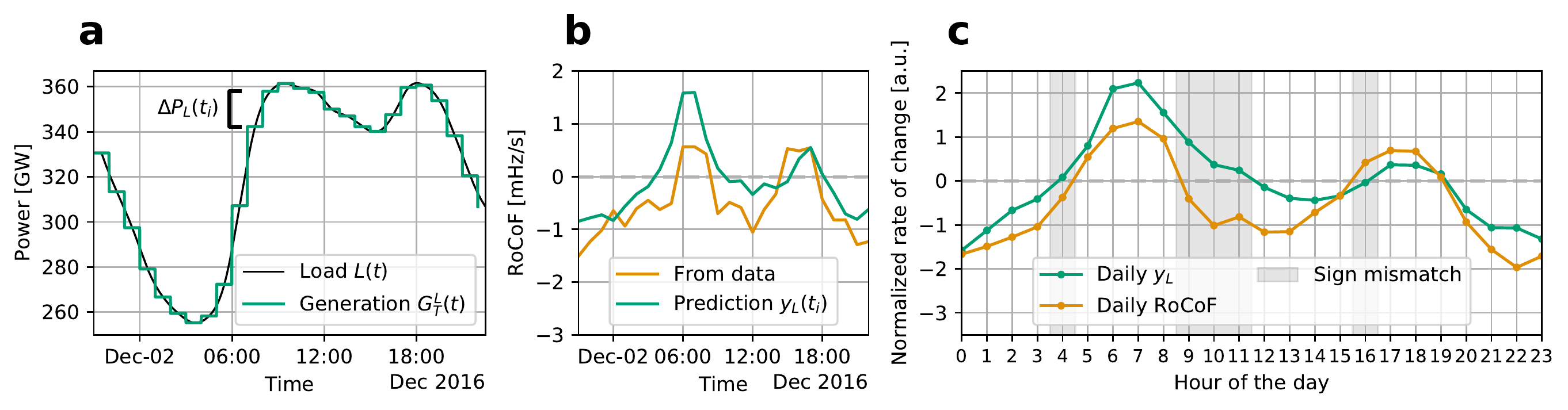}
    \caption{A common load-based model cannot reproduce the daily pattern of DFDs. a: A common model explains hourly DFDs with the power jump $\Delta P_L(t_i)$ that results from scheduled changes in the generation due to hourly trading on the electricity markets. b: Based on the generation steps $\Delta P_L$, we a estimate a linear model $y_L(t)$ for the hourly RoCoF, which we call the load-based model (eq. \ref{eq:1}). The prediction reproduces the data only partially and yields an R$^2$-score of 0.37 on the test set. The sample depicted in panel a and b is included in the test set. c: The load-based model cannot fully reproduce the daily pattern of DFDs, which is indicated by a sign mismatch between the daily average prediction (green) and the data-based RoCoF (orange).}
    \label{fig:2}
\end{figure*}

A common model for (hourly) DFDs is based on different temporal behaviour of the load and the generation, which is crucially determined by the electricity market \cite{weissbachHighFrequencyDeviations2009}. In Continental Europe, electricity is mostly traded in hourly blocks, such that power generators adapt their output to the new set point at the beginning of each hour. This leads to a step $\Delta P_L (t_i)$ in the power generation at the beginning of the hour $t_i$, as shown in fig.~\ref{fig:2}a. Since the load varies only slowly, the step-like generation introduces power imbalances around the full hour. For example, at 06:00 the imbalance changes from under- to oversupply due to the positive step $\Delta P_L$. The slope of the resulting frequency jump is then proportional to the generation step \cite{gorjaoDatadrivenModelPowergrid2019}, which implies a linear model $y_L(t_i)$ for the RoCoF:
\begin{align}
    y_L(t_i) = a \cdot \Delta P_L(t_i). 
    \label{eq:1}
\end{align}
Here, the value of the generation step $\Delta P_L(t)$ is estimated by the load ramp, i.e. the hourly change of the total load $\Delta L(t)$ in CE. This estimate is based on the assumption that most of the load is covered by schedule-based generation. The schedule coincides with the hourly average of the load, such that the generation steps approximate the load ramp. In that way, the direction of the DFDs is inherently correlated to the load ramp, so that we call the above approach the \textit{load-based} model.

To visualise the assumptions of the load-based model, we construct load and generation curves on sub-hourly resolution  (fig. \ref{fig:2}a). We design a continuous load curve $L(t)$ with a resolution of one minute by applying a cubic spline interpolation to the load values, which only have an hourly resolution (see sec. \ref{sec:data}). The generation values for each minute $G_T^L(t)$ follow a step function with hourly steps $\Delta P_L$. 

\subsection{Data sources and preparation} 
\label{sec:data}
To evaluate the load-based DFD model, we use a set of pre-processed, publicly available time series from the CE power system, which covers the years 2015 to 2019 ~\cite{kruseDataSetPredictionFrequency}. The data set contains hourly RoCoFs of the CE grid frequency, as well as multiple hourly-resolved \textit{external features}, such as the load and the power generation. 

\begin{figure}
    \centering
    \includegraphics[width=\columnwidth,clip,trim={0cm 1.8cm 0cm 0cm}]{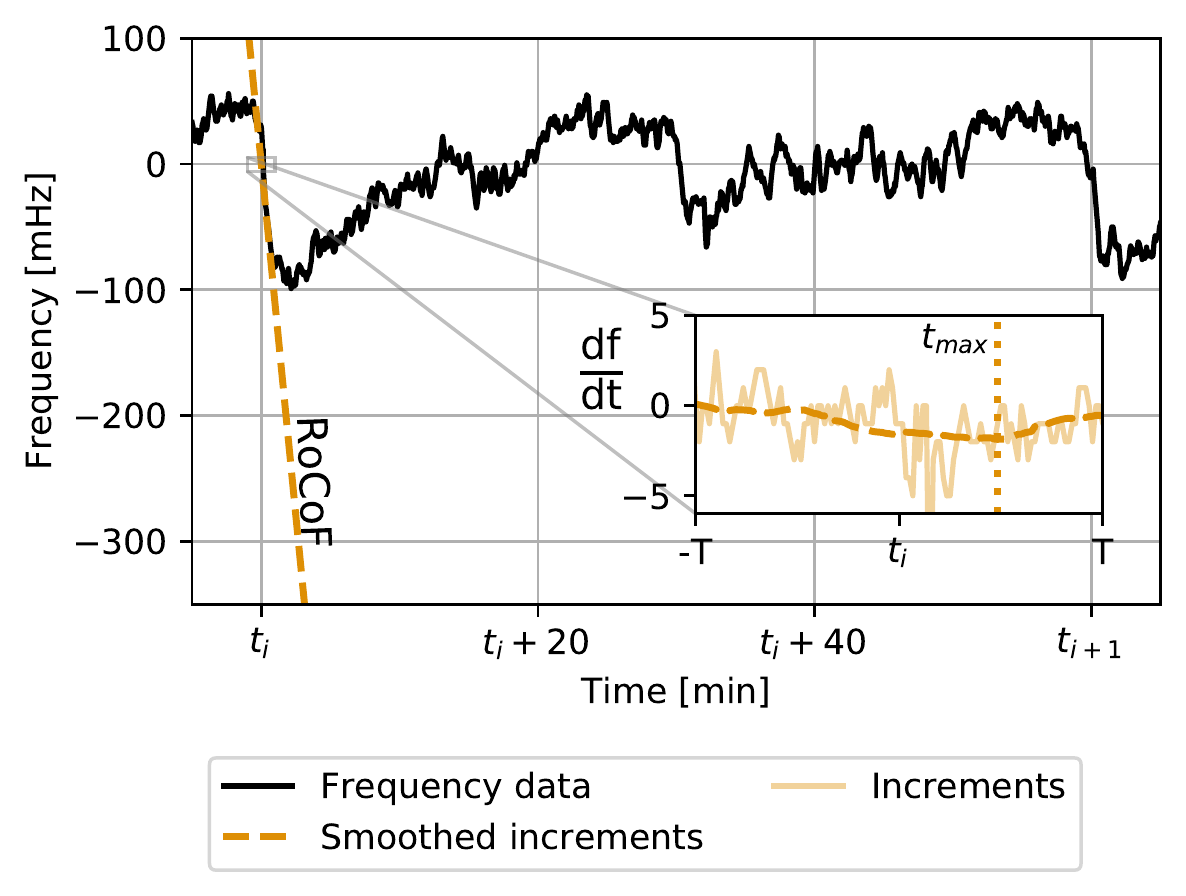}
    \caption{Quantifying DFDs with the hourly RoCoF. The grid frequency exhibits regular jumps, particularly at the beginning of the hour $t_i$, which are deterministic frequency deviations (DFDs). To measure the DFD at the full hour $t_i$, we calculate the corresponding Rate of Change of Frequency (RoCoF) from the grid frequency time series $f(t)$ (see inset).
    We first estimate the derivative $\textrm{d}f/\textrm{d}t$ by extracting and smoothing the increments $f(t+1\textrm{s})-f(t)$ (dashed orange line in the inset). We then evaluate the maximum slope in an interval around the full hour. See text for details. 
    \label{fig:1}
    }
\end{figure}

The hourly RoCof is extracted from grid frequency measurements $f(t)$ using the procedure sketched in fig.~\ref{fig:1}. Based on frequency data with 1 s resolution \cite{RegelenergieBedarfAbruf}, we estimate the derivative $\textrm{d} f/\textrm{d} t$ by first extracting the increments $f(t+ 1 s) -f(t)$. Then, we smooth the increments with a rectangular rolling window of length $L=30$ s. Finally, we identify the time $t_{max}$ in an interval around the full hour $[t_i-T,t_i+T]$, where the absolute derivative $|\textrm{d} f/\textrm{d} t|$ reaches its maximum. We then set $\textrm{RoCoF}(t_i) = \textrm{d} f/\textrm{d} t (t_{max})$ using $T=30 \textrm{s}$ (cf.~\cite{kruseRevealingRisks2021}).

The external features serve as input to the load-based model and to our ML model. From the data set, we only include a subset of features into our analysis. The subset includes the actual load and generation per type, as well as their hourly ramps and forecast errors. Moreover, we include day-ahead prices together with their ramps, and three variables indicating the hour, the weekday and the month. As our focus is on the explanation of DFDs and not on prediction, we here use the correct post-hoc values of all features and omit day-ahead predictions which are in principle available in the data set. As the grid frequency is affected by power imbalances in all locations of the synchronous area, the features represent the area-wide aggregated values of the variable. For example, the "load ramp" represents the slope of the aggregated load within CE. We refer to ref.~\cite{kruseRevealingRisks2021} for a detailed description of the data set, including the processing and the aggregation of the external features.

\subsection{Machine learning model for hourly RoCoF}
We apply the explainable ML model from ref. \cite{kruseRevealingRisks2021} to explore the impact of external features on the DFDs. The model uses a Gradient Tree Boosting model \cite{chenXGBoostScalableTree2016} for the prediction of hourly RoCoFs from external features, which is then explained through SHAP values. 

For the model training and evaluation, we randomly split the data set into a test set (20\%) and a training set (80\%). In contrast to ref.~\cite{kruseRevealingRisks2021}, we additionally include the data of a continuous 24h interval (from December 2016) in our test set to allow for a visualisation of the predicted time series (see fig. \ref{fig:2}b). Moreover, we only use the subset of features discussed in sec.~\ref{sec:data}. To optimise the model hyper-parameters, we use 5-fold cross-validation on the training set. Then, we retrain the optimal model on the whole training set and calculate the ML prediction $y_{ML}(t_i)$ of the RoCoF for every hour $t_i$ in the unseen test set. We evaluate the performance on the test set using the R$^2$-score, which represents the share of variability in the hourly RoCoF explained by the model. In the same way, we also estimate the linear model (eq.~\ref{eq:1}) on the training set (via least-squares) and evaluate it on the test set. In addition to the R$^2$-score, we evaluate the model predictions in terms of their daily average profile. A good model should reproduce the daily profile of RoCoFs estimated from the frequency data. 

To explain the model output, we finally calculate the SHAP values on the test set \cite{lundbergUnifiedApproachInterpreting2017}. SHAP values quantify the effect of each feature on the model prediction for each hour $t_i$. By aggregating individual SHAP values, we can also understand global relations in the model, in particular the dependency between external features, such as the load, and the hourly DFDs. In addition, we explore daily aggregated SHAP values that explain the daily average profile of the model prediction. They reflect the average impact of a feature on the RoCoF at a certain hour of the day and are therefore perfectly suited to explain the daily DFD pattern. For details on the ML model training, performance evaluation and model explanation via (daily aggregated) SHAP values we refer to ref. \cite{kruseRevealingRisks2021}.

\subsection{Data and code availability}
The data is publicly available on zenodo \cite{kruseDataSetPredictionFrequency} and the python code for this study can be obtained from GitHub \cite{frequency_rocof_repo}.

\begin{figure*}
    \centering
    \includegraphics[width=\textwidth]{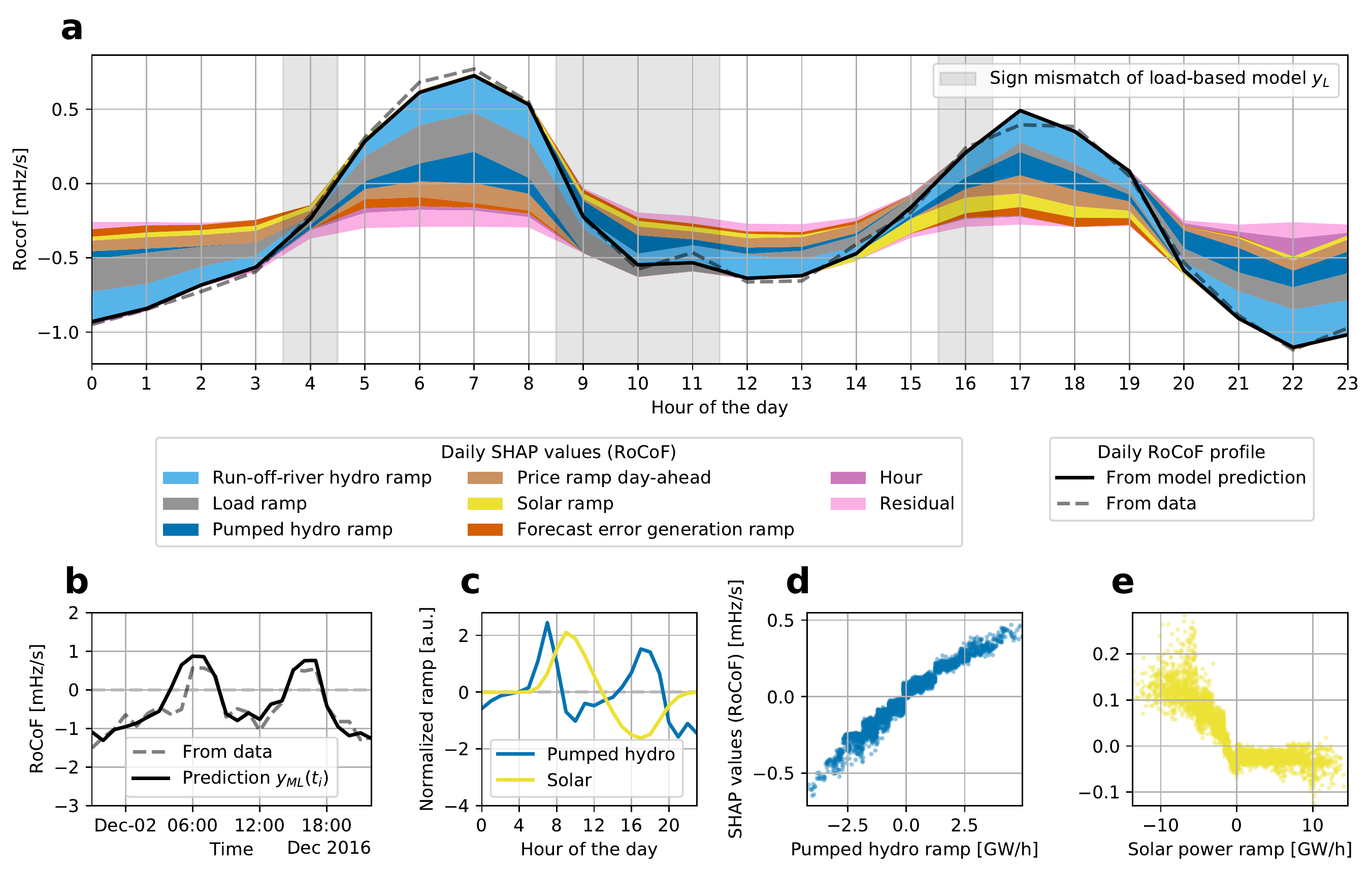}
    \caption{Revealing the effect of external features on daily DFDs with machine learning. a: We model the hourly RoCoF with an ML model based on external features, such as load ramps or forecast errors. By interpreting the model with SHAP, we can calculate daily (aggregated) SHAP values that explain the average impact of a feature on the model prediction for each hour of the day (coloured areas). For example, load ramps (grey) have a positive impact on the predicted RoCoF (solid line) at 07:00 and a negative impact at 22:00. b: The ML prediction $y_{ML}(t_i)$ predicts the hourly RoCoF with high accuracy in this sample, yielding an overall performance of R$^2=0.73$ on the test set. c: Driven by the SHAP results, we explain the sign mismatch of the load-based model with the daily ramps of solar power and fast generation, such as pumped hydro. d-e: Pumped hydro ramps have a positive relation to the RoCoF and solar ramps a non-linear dependency. The strong effect of negative pumped hydro ramps thus explains the deficits of the load-based model between 09:00 and 11:00 in the morning.}
    \label{fig:3}
\end{figure*}

\section{Results}
\label{sec:results}
\subsection{Evaluation of load-based DFD model}
The load-based predictor $y_L(t_i)$ (eq.~\ref{eq:1}) partially explains the direction of DFDs, but its overall performance is limited. Fig.~\ref{fig:2}b depicts the model prediction for a continuous time interval within the test set. The prediction reproduces the general trend of the RoCoF with upward jumps in the morning and in the afternoon, and downward jumps around noonday and during the night. However, the R$^2$-score only yields a value of 0.37, which reflects the mismatch of data and prediction in fig.~\ref{fig:2}b. 

Evaluating the daily profile of the prediction confirms this observation (fig.~\ref{fig:2}c). The overall shape of the daily RoCoF profile is reproduced, but the direction of the jumps does not align with the sign of the predicted RoCoF in 5 of 24 hours. In particular, the load is increasing on average between 09:00 and 11:00, but the average RoCoF is negative within these hours. We observe such a sign mismatch also at 04:00 and 16:00. The direction of the load ramp thus cannot fully explain the direction of the DFDs within the day. 

Note, that adding a constant bias to the linear model (eq.~\ref{eq:1}) increases the performance to R$^2=0.52$. However, the prediction is merely shifted and its daily profile still exhibits the wrong sign in 5 hours within the day. Consequently, an additional bias does not improve the explanation of the daily DFD pattern through the load-based model.

\subsection{Exploring DFDs with XAI}
\label{sec:shap}
We explain the deficits of the load-based model by exploring the daily DFDs with our explainable ML model (fig. \ref{fig:3}). As shown in fig.~\ref{fig:3}b, the ML prediction $y_{ML}(t_i)$ approximates the RoCoF in nearly every hour of the sample, which is consistent with the high R$^2$-score of 0.73 on the test set. We obtain an even higher correspondence for the daily profile (fig.~\ref{fig:3}a). The ML prediction (solid line) reproduces the daily RoCoF profile (dashed line) nearly perfectly. The model thus captures dependencies that are important to predict and explain the hourly RoCoF, and in particular the daily pattern of DFDs. 

Using our ML model explanation, daily aggregated SHAP values reveal the impact of external features on the daily DFD pattern (fig. \ref{fig:3}a). The SHAP values describe the contribution of each feature on the model prediction both in magnitude and sign. In the figure, the daily SHAP values are represented by coloured areas. Areas below the prediction line (solid) indicate a positive effect of the feature on the RoCoF in that hour, while areas below the line reflect a negative impact. We observe a positive impact of the load ramp on the RoCoF in the morning between 04:00 and 11:00, where the load is still rising (see fig. \ref{fig:2}c). This is in line with the load-based model, which predicts a positive RoCoF within this period due to the rising load. However, the actual RoCoF becomes negative between 09:00 and 11:00. Here, the contribution of the load ramp is still positive, but other features now have a strong impact on the model outcome. In fact, the ML model mainly attributes this decrease to the negative impact of hydro power ramps, which leads to a correct prediction of the downwards DFD direction.

\begin{figure*}
    \centering
    \includegraphics[width=\textwidth]{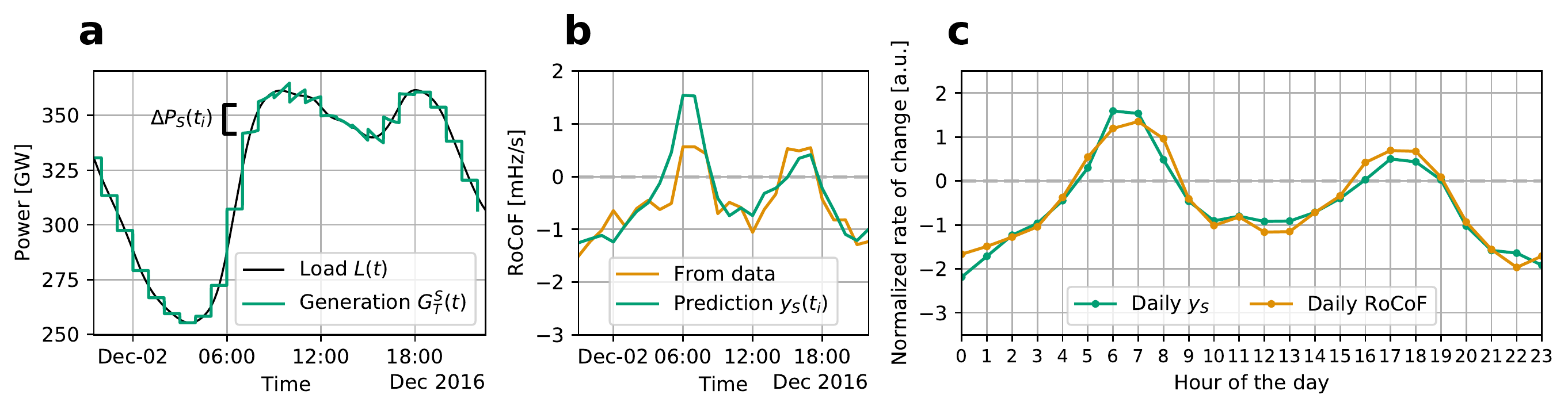}
    \caption{A refined physical model better reproduces daily DFDs. a: Based on our ML results, we calculate modified generation steps $\Delta P_S(t_i)$ that do not include continuous solar ramps. Instead, we interpolate the hourly solar power data to obtain a minute resolved solar generation $G_S(t)$. The panel sketches the new (total) generation $G_T^S(t)$ which combines the smooth solar generation and generation steps of size $\Delta P_S(t_i)$. b: By using a refined linear model $y_S(t_i)$ of the modified steps $\Delta P_S$ (eq. \ref{eq:2}), we improve the prediction of the load-based model. The refined model better reproduces the sample (orange line), with an overall performance of R$^2=0.63$ on the test set. c: The refined model also reproduces the direction of DFDs throughout the day.}
    \label{fig:4}
\end{figure*}

To further explain the contradiction between the ML prediction and the load-based model in the morning, we connect the ML prediction to the daily evolution of solar and hydro power (fig.~\ref{fig:3}c). Solar power ramps up slowly between 05:00 and 08:00. As the load already increases strongly during this time (see fig. \ref{fig:2}c), fast conventional generators, such as pumped hydro, ramp up to cover the load. Between 09:00 and 11:00 the load is still rising, but solar power starts to ramp up strongly and pumped hydro thus has to ramp down. As solar power ramps are still slow compared to hydro ramps, the generation step at the beginning of the hour is dominated by fast negative hydro ramps. Their direction is opposite to the load ramp, which is not captured in the load-based model. However, the ML model integrates this effect as depicted in the dependency plot fig.~\ref{fig:3}d, which displays the SHAP values together with the feature value. Negative hydro ramps have a strong negative impact on the predicted RoCoF. The ML model integrates this effect  (fig. \ref{fig:3}a) and thus reproduces the direction of the DFD correctly. 

In the afternoon, fast generation ramps and a smoothing effect of solar power explain the deficits of the load-based model. At 16:00 the load does not change strongly and the load-based model predicts a slightly negative RoCoF (see fig.~\ref{fig:2}c). However, solar power ramps down strongly in that time, so that flexible generators, such as pumped hydro, ramps up to cover the load (fig.~\ref{fig:3}c). The ML model captures this effect and the daily SHAP values in fig.~\ref{fig:3}a clearly show the positive effect of hydro power at 16:00, which leads to an upward DFD at this time. Interestingly, we also observe large positive SHAP values of solar ramps between 15:00 and 17:00. According to the dependency in fig.~\ref{fig:3}e, negative solar power ramps strongly increase the RoCoF, while positive ramps have a small negative effect. The impact of negative ramps most probably relates to a smoothing effect solar power. Solar power ramps down slowly thus following the load more smoothly than scheduled generation, which offsets the RoCoF. 

The sign mismatch of the load-based model at 04:00 is not explained by the ML model and probably relates to an overall bias in the hourly RoCoF. The ML prediction at 04:00 exhibits the average RoCoF value of $-0.29$ mHz/s (fig.~\ref{fig:3}a), which cannot be related to feature effects as SHAP values only explain the prediction \textit{relative} to the average. The average RoCoF reflects a bias of the hourly DFD towards negative values. For example, this bias can stem from different ramping speeds for upwards and downwards regulation of generators, or from a systematic bias in the power imbalance due to market mechanism. While our ML model reproduces the bias of the average RoCoF, a physical explanation thereof is still open. 

Finally, we point to the impact of local legislation that is revealed by the "hour" feature in fig.~\ref{fig:3}a. The hour exhibits its only (strong) effect at 22:00. At this time wind power farms in Germany are shut down due to environmental protection in favour of bats and noise prevention \cite{entso-eaisblReportDeterministicFrequency2019}. This leads to a strong negative effect on the RoCoF, which is well captured in our ML model, but not included in the load-based model at all.

\subsection{A refined view of DFDs}
The ML results lead us to a refined view of the generation behaviour that cause hourly DFDs (fig. \ref{fig:4}a). To include the continuous effect of solar power into the load-based model, we calculate modified generation steps by subtracting the hourly solar power ramp $\Delta G_S(t_i)$:
\begin{align}
\Delta P_S(t_i) = \Delta P_L(t_i) - \Delta G_S(t_i),    
\end{align}
In this way, we define a refined linear model that predicts the RoCoF based on the power jump $\Delta P_S(t_i)$:
\begin{align}
    y_S(t_i)=a\cdot \Delta P_S(t_i)+b.
    \label{eq:2}
\end{align}
Here, we introduced the intercept $b$ to model the negative bias of the RoCoF as discussed in sec. \ref{sec:shap}. 

To visualise the refined model, we calculate the solar generation $G_S(t)$ for every minute with a cubic spline interpolation of the hourly solar power time series in our data set. The refined generation curve is then given by $G_T^S(t) =  G_L(t) + G_S(t)$, where $G_L$ has hourly steps $\Delta P_S$. The resulting sketch in fig.~\ref{fig:4}a depicts the same sample as in fig.~\ref{fig:2}a, but the generation steps have another direction particularly at 09:00. We clearly see that the continuous solar power ramps up faster than the load between 09:00 and 10:00. As explained through our ML model, this leads to fast downwards ramps of flexible generation, as observed in fig. \ref{fig:4}a at 09:00. Notably, we can also observe the smoothing effect of solar power between 13:00 and 15:00. Here, the solar generation smoothly follows the load thus leading to a less severe generation step than predicted through the load-based model. 

The refined model of hourly power steps outperforms the load-based model and explains the daily DFD pattern successfully. The prediction of the continuous sample in fig.~\ref{fig:4}b better aligns with the data, as compared to the load-based model, and the overall R$^2$-score yields 0.63 on the test set. Furthermore, the refined model correctly reproduces the direction of the daily RoCoFs in each hour of the day (fig.~\ref{fig:4}c). Together with a model bias, the approximation of hourly generation ramp \textit{without} the contribution of solar power already leads to a good explanation of the daily DFD pattern. 

\section{Conclusion}
In summary, we have modelled and explored the relation between DFDs and external features such as load and generation ramps using explainable ML. Our ML model explains 73\% of the variation within the slope of hourly DFDs. The evaluation of a common, load-based DFD model yields a much lower performance of 37\% and does not entirely explain the daily pattern of DFDs in Continental Europe. Using daily aggregated SHAP values, we offer another way to explain the daily DFD pattern, thus revealing multiple important external features that go beyond a pure load-based view. Including this information into the load-based predictor improves its performance to 63\% and gives a refined view on daily DFD patterns. 

Our explainable ML model reveals that solar ramps and scheduled (fast) generation ramps are important for the daily DFD pattern. Solar power ramps continuously during the day. Meanwhile, fast scheduled generation ramps, from e.g. hydro power, have to balance this behaviour thus leading to DFDs that can be opposite to the slope of the load curve. This is consistent with other studies that point to the connection between solar power and generation ramps on 15 min basis due to intraday electricity trading 
\cite{kochShorttermElectricityTrading2019, weissbachImpactCurrentMarket2018}.  

Our explainable ML approach provides an alternative to physical models of the grid frequency, but also complements them. Physical models such as the DFD model in ref.~\cite{weissbachHighFrequencyDeviations2009} have limits when modelling the impact of multiple external features on frequency stability for each hour. Control parameters and highly resolved generation data are mostly not publicly available, such that simulation have to use simplifying assumptions. Our ML model therefore offers an alternative by using publicly available data and estimating a model for DFDs without physical simplifications. However, the ML approach can also complement physical models by pointing to unknown dependencies and important features that have to be included to improve physical modelling. For example, we have presented a refined model of hourly generation steps using our ML insights, which improves the common load-based view. 

As frequency deviations have been increasing in Continental Europe, a limitation of large DFDs is urgently needed \cite{entso-eaisblReportDeterministicFrequency2019}. Data-based predictions can be combined with control actions to prevent large DFDs \cite{avramiotis-falireasMPCStrategyAutomatic2013}, thus saving costs for control actions and making the grid less vulnerable. Explainable ML models can offer predictions \cite{kruseRevealingRisks2021}, but they also contribute by better understanding drivers and risks of DFDs in an increasingly complex power system.

\bibliographystyle{IEEEtran}
\bibliography{IEEEabrv,references}

\begin{thebibliography}{10}
\providecommand{\url}[1]{#1}
\csname url@samestyle\endcsname
\providecommand{\newblock}{\relax}
\providecommand{\bibinfo}[2]{#2}
\providecommand{\BIBentrySTDinterwordspacing}{\spaceskip=0pt\relax}
\providecommand{\BIBentryALTinterwordstretchfactor}{4}
\providecommand{\BIBentryALTinterwordspacing}{\spaceskip=\fontdimen2\font plus
\BIBentryALTinterwordstretchfactor\fontdimen3\font minus
  \fontdimen4\font\relax}
\providecommand{\BIBforeignlanguage}[2]{{%
\expandafter\ifx\csname l@#1\endcsname\relax
\typeout{** WARNING: IEEEtran.bst: No hyphenation pattern has been}%
\typeout{** loaded for the language `#1'. Using the pattern for}%
\typeout{** the default language instead.}%
\else
\language=\csname l@#1\endcsname
\fi
#2}}
\providecommand{\BIBdecl}{\relax}
\BIBdecl

\bibitem{machowskiPowerSystemDynamics2008}
J.~Machowski, J.~Bialek, J.~Bumby, and D.~J. Bumby, \emph{Power {{System
  Dynamics}}: {{Stability}} and {{Control}}}.\hskip 1em plus 0.5em minus
  0.4em\relax {New York}: {John Wiley \& Sons, Ltd.}, 2008.

\bibitem{schaferNonGaussianPowerGrid2018}
B.~Sch{\"a}fer, C.~Beck, K.~Aihara, D.~Witthaut, and M.~Timme,
  ``\BIBforeignlanguage{en}{Non-{{Gaussian}} power grid frequency fluctuations
  characterized by {{L\'evy}}-stable laws and superstatistics},''
  \emph{\BIBforeignlanguage{en}{Nature Energy}}, vol.~3, no.~2, pp. 119--126,
  Feb. 2018.

\bibitem{homanGridFrequencyVolatility2021}
S.~Homan, N.~Mac~Dowell, and S.~Brown, ``\BIBforeignlanguage{en}{Grid frequency
  volatility in future low inertia scenarios: {{Challenges}} and mitigation
  options},'' \emph{\BIBforeignlanguage{en}{Applied Energy}}, vol. 290, p.
  116723, May 2021.

\bibitem{liFrequencyDeviationsGeneration2011}
Z.~W. Li, O.~Samuelsson, and R.~{Garcia-Valle}, ``Frequency deviations and
  generation scheduling in the nordic system,'' in \emph{2011 {{IEEE Trondheim
  PowerTech}}}, Jun. 2011, pp. 1--6.

\bibitem{entso-eaisblReportDeterministicFrequency2019}
{ENTSO-E}, ``\BIBforeignlanguage{en}{Report on {{Deterministic Frequency
  Deviations}}},''
  \url{https://consultations.entsoe.eu/system-development/deterministic\_frequency\_deviations\_report/user\_uploads/report\_deterministic\_frequency\_deviations\_final-draft-for-consultation.pdf},
  2019.

\bibitem{weissbachHighFrequencyDeviations2009}
T.~Weissbach and E.~Welfonder, ``High frequency deviations within the
  {{European Power System}}: {{Origins}} and proposals for improvement,'' in
  \emph{2009 {{IEEE}}/{{PES Power Systems Conference}} and
  {{Exposition}}}.\hskip 1em plus 0.5em minus 0.4em\relax {Seattle}: {IEEE},
  Mar. 2009, pp. 1--6.

\bibitem{gorjaoDatadrivenModelPowergrid2019}
L.~R. Gorj{\~a}o, M.~Anvari, H.~Kantz, C.~Beck, D.~Witthaut, M.~Timme, and
  B.~Sch{\"a}fer, ``Data-driven model of the power-grid frequency dynamics,''
  \emph{IEEE Access}, vol.~8, pp. 43\,082--43\,097, 2020.

\bibitem{perssonFrequencyEvaluationNordic2017}
M.~Persson and P.~Chen, ``Frequency evaluation of the {{Nordic}} power system
  using {{PMU}} measurements,'' \emph{Transmission Distribution IET
  Generation}, vol.~11, no.~11, pp. 2879--2887, 2017.

\bibitem{kruseRevealingRisks2021}
J.~Kruse, B.~Sch\"afer, and D.~Witthaut, ``Revealing drivers and risks for
  power grid frequency stability with explainable {AI},'' Preprint at
  \url{https://arxiv.org/abs/2106.04341}, 2021.

\bibitem{krusePredictabilityPowerGrid2020}
J.~Kruse, B.~Sch{\"a}fer, and D.~Witthaut, ``Predictability of {{Power Grid
  Frequency}},'' \emph{IEEE Access}, vol.~8, pp. 149\,435--149\,446, 2020.

\bibitem{hastieElementsStatisticalLearning2016}
T.~Hastie, R.~Tibshirani, and J.~Friedman,
  \emph{\BIBforeignlanguage{English}{The {{Elements}} of {{Statistical
  Learning}}: {{Data Mining}}, {{Inference}}, and {{Prediction}}}},
  2nd~ed.\hskip 1em plus 0.5em minus 0.4em\relax {New York}: {Springer}, 2016.

\bibitem{hirthENTSOETransparencyPlatform2018}
L.~Hirth, J.~M{\"u}hlenpfordt, and M.~Bulkeley, ``The {{ENTSO}}-{{E
  Transparency Platform}} \textendash{} {{A}} review of {{Europe}}'s most
  ambitious electricity data platform,'' \emph{Applied Energy}, vol. 225, pp.
  1054--1067, Sep. 2018.

\bibitem{rydingorjaoOpenDatabaseAnalysis2020}
L.~Rydin~Gorj{\~a}o, R.~Jumar, H.~Maass, V.~Hagenmeyer, G.~C. Yalcin, J.~Kruse,
  M.~Timme, C.~Beck, D.~Witthaut, and B.~Sch{\"a}fer,
  ``\BIBforeignlanguage{en}{Open database analysis of scaling and
  spatio-temporal properties of power grid frequencies},''
  \emph{\BIBforeignlanguage{en}{Nature Communications}}, vol.~11, no.~1, p.
  6362, Dec. 2020.

\bibitem{roscherExplainableMachineLearning2020}
R.~Roscher, B.~Bohn, M.~F. Duarte, and J.~Garcke, ``Explainable {{Machine
  Learning}} for {{Scientific Insights}} and {{Discoveries}},'' \emph{IEEE
  Access}, vol.~8, pp. 42\,200--42\,216, 2020.

\bibitem{barredoarrietaExplainableArtificialIntelligence2020}
A.~Barredo~Arrieta, N.~{D{\'i}az-Rodr{\'i}guez}, J.~Del~Ser, A.~Bennetot,
  S.~Tabik, A.~Barbado, S.~Garcia, S.~{Gil-Lopez}, D.~Molina, R.~Benjamins,
  R.~Chatila, and F.~Herrera, ``\BIBforeignlanguage{en}{Explainable
  {{Artificial Intelligence}} ({{XAI}}): {{Concepts}}, taxonomies,
  opportunities and challenges toward responsible {{AI}}},''
  \emph{\BIBforeignlanguage{en}{Information Fusion}}, vol.~58, pp. 82--115,
  Jun. 2020.

\bibitem{lundbergLocalExplanationsGlobal2020a}
S.~M. Lundberg, G.~Erion, H.~Chen, A.~DeGrave, J.~M. Prutkin, B.~Nair, R.~Katz,
  J.~Himmelfarb, N.~Bansal, and S.-I. Lee, ``\BIBforeignlanguage{en}{From local
  explanations to global understanding with explainable {{AI}} for trees},''
  \emph{\BIBforeignlanguage{en}{Nature Machine Intelligence}}, vol.~2, no.~1,
  pp. 56--67, Jan. 2020.

\bibitem{lundbergUnifiedApproachInterpreting2017}
S.~M. Lundberg and S.-I. Lee, ``A unified approach to interpreting model
  predictions,'' in \emph{Proceedings of the 31st {{International Conference}}
  on {{Neural Information Processing Systems}}}, ser. {{NIPS}}'17.\hskip 1em
  plus 0.5em minus 0.4em\relax {New York}: {Curran Associates Inc.}, Dec. 2017,
  pp. 4768--4777.

\bibitem{lundbergConsistentIndividualizedFeature2019}
S.~M. Lundberg, G.~G. Erion, and S.-I. Lee, ``Consistent {{Individualized
  Feature Attribution}} for {{Tree Ensembles}},'' Preprint at
  \url{https://arxiv.org/abs/1802.03888}, 2019.

\bibitem{weissbachImpactCurrentMarket2018}
T.~Wei{\ss}bach, S.~Remppis, and H.~Lens, ``Impact of {{Current Market
  Developments}} in {{Europe}} on {{Deterministic Grid Frequency Deviations}}
  and {{Frequency Restauration Reserve Demand}},'' in \emph{2018 15th
  {{International Conference}} on the {{European Energy Market}} ({{EEM}})},
  Jun. 2018, pp. 1--6.

\bibitem{kruseDataSetPredictionFrequency}
J.~Kruse, B.~Sch{\"a}fer, and D.~Witthaut, ``Data set for predicting grid
  frequency stability indicators,'' The data will be publicly available upon
  publication.

\bibitem{RegelenergieBedarfAbruf}
{TransnetBW GmbH}, ``\BIBforeignlanguage{de}{{Regelenergie Bedarf + Abruf}},''
  \url{https://www.transnetbw.de/de/strommarkt/systemdienstleistungen/regelenergie-bedarf-und-abruf},
  2020.

\bibitem{chenXGBoostScalableTree2016}
T.~Chen and C.~Guestrin, ``{{XGBoost}}: {{A Scalable Tree Boosting System}},''
  in \emph{Proceedings of the 22nd {{ACM SIGKDD International Conference}} on
  {{Knowledge Discovery}} and {{Data Mining}}}, ser. {{KDD}} '16.\hskip 1em
  plus 0.5em minus 0.4em\relax {New York}: {ACM}, Aug. 2016, pp. 785--794.

\bibitem{frequency_rocof_repo}
J.~Kruse, B.~Sch{\"a}fer, and D.~Witthaut, ``An explainable model for
  deterministic grid frequency deviations in continental europe,'' The code
  will be available on github upon publication.

\bibitem{kochShorttermElectricityTrading2019}
C.~Koch and L.~Hirth, ``\BIBforeignlanguage{en}{Short-term electricity trading
  for system balancing: {{An}} empirical analysis of the role of intraday
  trading in balancing {{Germany}}'s electricity system},''
  \emph{\BIBforeignlanguage{en}{Renewable and Sustainable Energy Reviews}},
  vol. 113, p. 109275, Oct. 2019.

\bibitem{avramiotis-falireasMPCStrategyAutomatic2013}
I.~{Avramiotis-Falireas}, A.~Troupakis, F.~Abbaspourtorbati, and M.~Zima, ``An
  {{MPC}} strategy for automatic generation control with consideration of
  deterministic power imbalances,'' in \emph{2013 {{IREP Symposium Bulk Power
  System Dynamics}} and {{Control}} - {{IX Optimization}}, {{Security}} and
  {{Control}} of the {{Emerging Power Grid}}}, Aug. 2013, pp. 1--8.

\end{thebibliography}

\end{document}